\documentclass{article} % For LaTeX2e
\usepackage{iclr2026_conference,times}
\usepackage{amsmath,amsfonts,bm}

\def\eqref#1{equation~\ref{#1}}
\def\1{\bm{1}}

\DeclareMathAlphabet{\mathsfit}{\encodingdefault}{\sfdefault}{m}{sl}
\SetMathAlphabet{\mathsfit}{bold}{\encodingdefault}{\sfdefault}{bx}{n}

\usepackage{hyperref}
\usepackage{url}
\usepackage{float}
\usepackage{booktabs}
\usepackage{multirow}
\usepackage{graphicx}
\usepackage{algorithm}
\usepackage{algpseudocode}
\title{MIND: Multi-agent Inference for Negotiation Dialogue in Travel Planning}

\iclrfinalcopy % Camera-ready version 활성화

\author{
  \textbf{Hunmin Do}$^1$ \quad \textbf{Taejun Yoon}$^2$ \quad \textbf{Kiyong Jung}$^3$ \\
  $^1$School of Mechanical Engineering, Sungkyunkwan University, Suwon, Korea \\
  \texttt{gnsals9262@g.skku.edu} \\
  $^2$Department of Applied Artificial Intelligence, Sungkyunkwan University, Seoul, Korea \\
  \texttt{ohimfrog03@g.skku.edu} \\
  $^3$Department of Software, Sungkyunkwan University, Suwon, Korea \\
  \texttt{wjdrldyd0213@g.skku.edu}
}

\begin{document}

\maketitle

\begin{abstract}
While Multi-Agent Debate (MAD) research has advanced, its efficacy in coordinating complex stakeholder interests---such as travel planning---remains largely unexplored. To bridge this gap, we propose MIND (Multi-agent Inference for Negotiation Dialogue), a framework designed to simulate realistic consensus-building among travelers with heterogeneous preferences. Grounded in the Theory of Mind (ToM), MIND introduces a Strategic Appraisal phase that infers opponent willingness ($w$) from linguistic nuances with 90.2\% accuracy. Experimental results demonstrate that MIND outperforms traditional MAD frameworks, achieving a 20.5\% improvement in High-$w$ Hit and a 30.7\% increase in Debate Hit-Rate, effectively prioritizing high-stakes constraints. Furthermore, qualitative evaluations via LLM-as-a-Judge confirm that MIND surpasses baselines in Rationality (68.8\%) and Fluency (72.4\%), securing an overall win rate of 68.3\%. These findings validate that MIND effectively models human negotiation dynamics to derive persuasive consensus.
\end{abstract}
\section{Introduction}
Recently, research on Multi-Agent Debate (MAD) \citep{du2024improving} utilizing Large Language Models (LLMs) has emerged as a pivotal paradigm for overcoming the limitations of individual models and eliciting collective intelligence. While conventional MAD studies have predominantly focused on tasks with explicit ground truths, such as mathematics or coding, recent initiatives like \textit{Debate-to-Write} \citep{hu-etal-2025-debate} have attempted to secure diversity of thought and logical consistency in subjective argumentation through persona-based debates. In practice, real-world decision-making is often closer to a ``social cognitive process''—reconciling divergent perspectives and subjective preferences to reach a consensus—rather than a search for a single, fixed answer.

While existing research in travel planning \citep{xie2024travelplanner, chaudhuri-etal-2025-tripcraft, shao-etal-2025-personal} has successfully addressed complex constraints, these approaches remain largely confined to single-agent optimization problems. To bridge this gap, this study proposes \textbf{MIND (Multi-agent Inference for Negotiation Dialogue)}, which extends the domain of travel planning into a social decision-making process necessitating multi-party compromise. By enabling each persona to recognize its own preference intensity and engage in strategic communication, we establish a dynamic negotiation framework grounded in cognitive principles rather than mere information aggregation.

\section{Related Works}

\subsection{LLM-based Travel Planning}
Benchmarks such as TravelPlanner \citep{xie2024travelplanner} and TripCraft \citep{chaudhuri-etal-2025-tripcraft} established the foundation for complex reasoning in travel. While recent systems such as TripTailor \citep{wang-etal-2025-triptailor} and Personal Travel Solver \citep{shao-etal-2025-personal}, proposed systems that optimize individual preferences by integrating mathematical solvers with LLMs. However, these works predominantly treat travel planning as a single-persona optimization problem, often simplifying travel companions into static numerical variables. Consequently, they fail to capture the social dynamics—specifically, negotiation and compromise—that are central to real-world group travel, where conflicting preferences must be reconciled.

\subsection{Multi-Agent Debate (MAD) and Social Cognition}
Recent research in Multi-Agent Debate (MAD) has drawn inspiration from the “Society of Mind” paradigm \citep{societyofmind}, exploring social phenomena where consensus emerges through agent interactions. Strategic behavior in negotiation dialogues, however, remains a significant challenge for AI \citep{lewis-etal-2017-deal}. Building on these foundations, simulating Theory of Mind (ToM) tasks through task decomposition \citep{sarangi-etal-2025-decompose} has provided a cognitive basis for agents to infer others' internal states.While existing efforts to improve MAD mechanisms \citep{kaesberg-etal-2025-voting,pitre-etal-2025-consensagent} largely focus on objective tasks with explicit ground truths, research remains sparse regarding subjective scenarios with conflicting preferences. Recently, the Debate-to-Write framework \citep{hu-etal-2025-debate} demonstrated that persona-based debates can enhance diversity and consistency in subjective argumentation, extending MAD's boundaries.Despite these advancements, the negotiation dynamics emphasized by the Dual Concern Model \citep{carnevale1992negotiation}—specifically the tension between self-interest and concern for others—remain under-explored in information-asymmetric environments. This study addresses this gap by proposing the MIND framework, which integrates Strategic Appraisal for intent analysis and dynamic tone adjustment regulated by preference intensity ($w$).

\section{Methodology}

\subsection{Multi-Persona Data Augmentation}To simulate realistic group dynamics, we enhanced the \textit{TravelPlanner} \citep{xie2024travelplanner} benchmark by integrating preference-rich attributes from the \textit{Stravl} \citep{stravl2023data} dataset. The augmentation followed a three-stage pipeline. First, we extracted 200–400 distinct candidate personas per scenario using the MMR (Maximum Marginal Relevance) algorithm to ensure diversity. Second, we synthesized context-aware responses to 20 additional survey questions from \textit{Stravl} using an LLM. Adopting the MoSCoW prioritization framework, we further derived a Willingness score ($w \in [1, 10]$) for each preference to quantify its strategic importance. Third, we applied a filtering protocol to form final groups based on the following constraint definitions. \textbf{Hard Constraints} refer to non-negotiable requirements essential for the basic viability of a trip—such as matching travel dates or departure locations—which must be shared by all members without conflict to ensure the plan's execution. In contrast, \textbf{Soft Constraints} are defined as subjective preferences with high strategic importance ($6 \le w \le 8$) where members possess divergent needs. We specifically formed groups with at least three such conflicts to deliberately foster a competitive negotiation environment that necessitates substantive compromise rather than trivial consensus.

\subsection{MIND (Multi-agent Inference for Negotiation Dialogue)}
We designed the MIND framework to analyze the process by which multiple participants reach an agreement while considering mutual satisfaction. Specifically, we focus on an environment of \textbf{information asymmetry (Hidden-$w$)}, where agents know only their own $w$ values and not those of their counterparts.
The primary independent variable is the presence of \textbf{Linguistic Nuance Injection}. In the Base Discussion, agents negotiate without explicit tone changes based on $w$. In contrast, the MIND enables agents to dynamically adjust their linguistic tone between warmth and toughness proportional to $w$.

Our agents perform advanced social reasoning based on Task Decomposition \citep{sarangi-etal-2025-decompose}. Upon receiving a proposal, an agent undergoes a Strategic Appraisal phase:
\begin{enumerate}
    \item \textbf{Inference:} Analyze the opponent's tone and argument strength to infer their hidden $w$ (\textit{Guessed Opponent $w$}).
    \item \textbf{Decision:} Determine a Strategy Intent (Push, Compromise, or Yield) by comparing the guessed $w$ with their own.
\end{enumerate}
If a consensus is not reached via a majority vote within three rounds, a Fallback mechanism is applied, adopting the opinion of the agent with the highest $w$ to prevent the collapse of overall group utility.

\begin{figure*}[t!]
    \centering
    \includegraphics[width=1.0\linewidth]{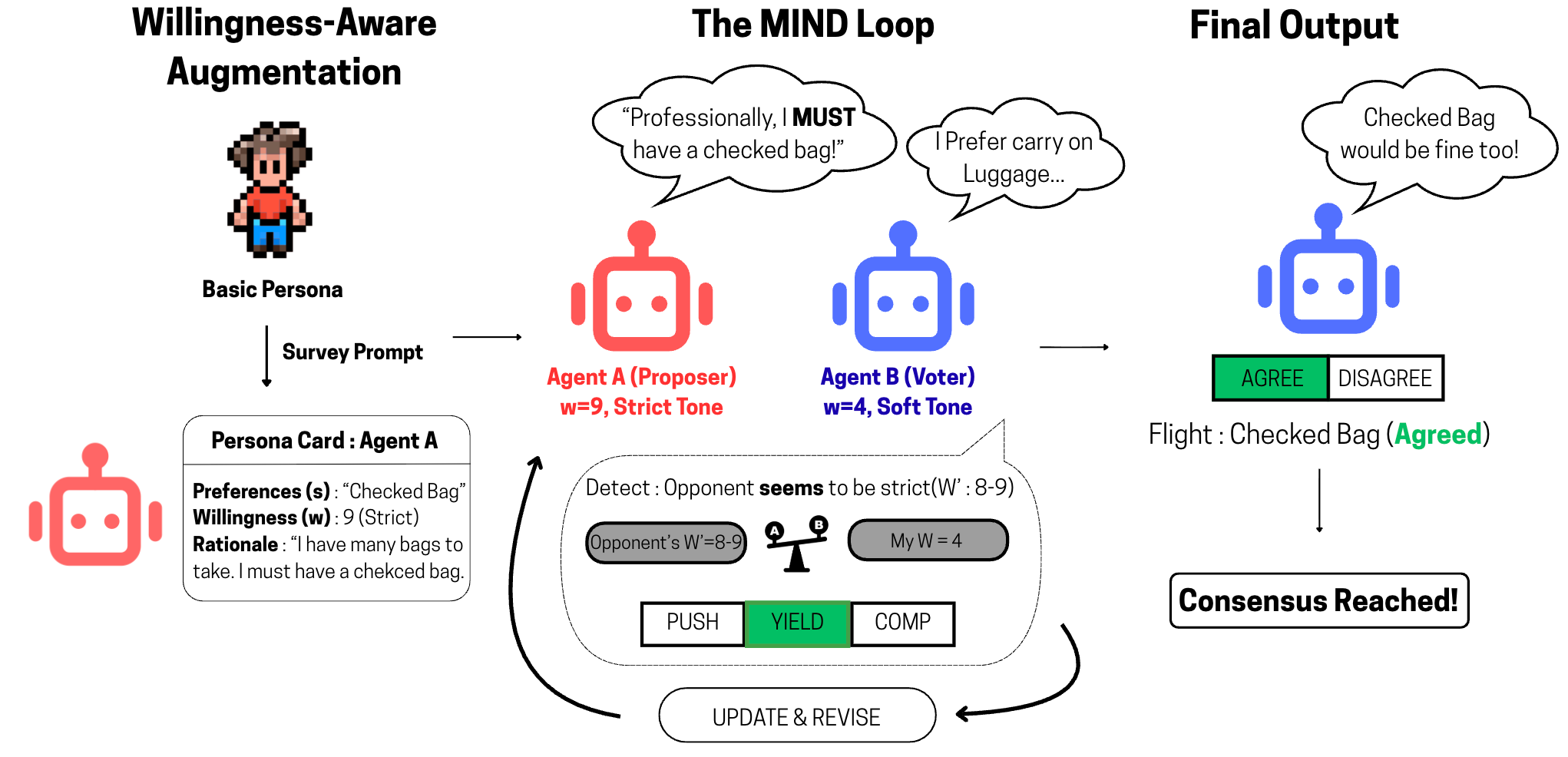}
    \vspace{-13pt} % 이 부분을 추가해서 간격을 줄여봐! (값은 조절 가능)
    \caption{\textbf{Overview of the MIND Framework.} (Left) Persona profiles are augmented with Willingness ($w$) values. (Center) Agents engage in the \textit{MIND Loop}, inferring opponents' hidden $w$ and adjusting strategies. (Right) The process yields a consensus travel plan.}
    \label{fig:mind_flowchart}
\end{figure*}
\section{Experiments and Results}

\subsection{Experimental Setup \& Metrics}
All experiments utilize \texttt{gpt-4.1-mini-2025-04-14} (Temp=0.4) as the backbone agent. To reflect the most common social units in travel—ranging from couples to nuclear families—we utilized the augmented persona data to vary the group size from 2 to 4 agents for each scenario. We define five metrics to evaluate group decision-making across 201 negotiation scenarios, where $v_{i,c}$ is the initial preference, $V_c$ is the final agreement, and $A$ is the agent set. Here, $C$ denotes the set of negotiation cases ($|C| = 201$), and $c \in C$ indexes a specific negotiation case.

\begin{itemize}
    \item \textbf{Total Fidelity ($F$):} The average proportion of individual preferences preserved in the final agreement across all participants.
    \begin{equation}
        F = \frac{1}{|A| \cdot |C|} \sum_{c \in C} \sum_{i \in A} \mathbf{1}(v_{i,c} = V_c)
    \end{equation}

    \item \textbf{Debate Hit-Rate (DHR)}: Specifically measures whether the High-$w$ agent's opinion prevailed within voluntary debates ($C_{debate}$), indicating strategic efficiency. 
    \begin{equation}
    DHR = \frac{1}{|C_{debate}|} \sum_{c \in C_{debate}} \mathbf{1}(\exists i \in Top(c) : v_{i,c} = V_c)
    \end{equation}

    \item \textbf{Debate Ratio (DR)}: The ratio of total negotiation items where a voluntary agreement was reached through agent deliberation without resorting to forced fallback mechanisms.
    \begin{equation}
    DR = \frac{|C_{debate}|}{|C|}
    \end{equation}

    \item \textbf{Total Satisfaction ($S_{total}$)}: The sum of weighted satisfaction scores of all agents in the group, representing the overall social welfare.
    \begin{equation}
    S_{total} = \sum_{i \in A} \sum_{c \in C} (w_{i,c} \cdot \mathbf{1}(v_{i,c} = V_c))
    \end{equation}
    
    \item \textbf{Fairness ($\mathcal{J}$)}: We use Jain's Fairness Index \citep{jain1984quantitative} to measure the distributional equity of the weighted satisfaction sum $S_i$ per agent, defined as $S_i = \sum_{c \in C} (w_{i,c} \cdot \mathbb{1}(v_{i,c} = V_c))$. A value closer to 1 indicates that satisfaction is distributed fairly across the group.
    \begin{equation}
    \mathcal{J} = \frac{(\sum_{i \in A} S_i)^2}{|A| \cdot \sum_{i \in A} S_i^2}
    \end{equation}
\end{itemize}

\paragraph{ToM Inference Accuracy.} To evaluate the cognitive foundation of the Strategic Appraisal phase, we measure the error between the inferred willingness ($w_{pred}$) and the ground truth ($w_{true}$).
    \begin{itemize}
        \item \textbf{Mean Absolute Error (MAE)}: $\frac{1}{N} \sum |w_{true} - w_{pred}|$, measuring the average magnitude of estimation errors.
        \item \textbf{Accuracy within $\pm\delta$}: The proportion of inferences where $|w_{true} - w_{pred}| \le \delta$. We report for $\delta=1$ and $\delta=2$ to assess the model's proximity to actual intent.
        \item \textbf{Pearson Correlation ($r$)}: Measures the linear relationship between true and predicted $w$ to evaluate the model's ability to capture willingness trends.
    \end{itemize}

\paragraph{item Qualitative Evaluation (LLM-as a-Judge).} We utilize \texttt{gpt-4.1-2025-04-14} to evaluate the linguistic and strategic quality of dialogs in three dimensions: Rationality (logical consistency) and Fluency (naturalness).

\subsection{Results and Analysis}

\paragraph{Quantitative Performance \& Strategic Trade-off.}
As shown in Table~\ref{tab:quantitative}, the MIND demonstrates significant strategic superiority, recording 35.08\% in High-$w$ Hit (+20.5\%) and 34.65\% in Debate Hit-Rate (+30.7\%). Notably, the Debate Ratio reached 93.18\%, confirming that agreements were reached through substantial deliberation. 
The High-$w$ Hit increase validates our Willingness-Weighted Efficiency. Unlike mechanical averaging, MIND agents yield low-priority items ($w \le 4$) to secure high-priority constraints ($w \ge 8$), avoiding the ``tyranny of the average'' by prioritizing essential needs through strategic deliberation.

\begin{table}[h]
    \centering
    \caption{Performance Comparison ($N=201$). MIND shows superior strategic efficiency.}
    \label{tab:quantitative}
    \resizebox{\textwidth}{!}{
    \begin{tabular}{lcccccc}
        \toprule
        \textbf{Method} & \textbf{Debate Hit-Rate} & \textbf{Debate Ratio} & \textbf{Fairness} & \textbf{Total Fidelity} & \textbf{Total Sat. ($S_{total}$)} \\
        \midrule
        Base & 26.51\% & 82.71\% & 0.6849 & 25.80\% & 18.03 \\
        \textbf{MIND} & \textbf{34.65\%} & \textbf{93.18\%} & 0.6838 & 23.87\% & \textbf{19.96} \\
        \bottomrule
    \end{tabular}
    }
\end{table}

\paragraph{Scalability Analysis.}
Table~\ref{tab:scalability} illustrates the robustness of MIND across varying group sizes (2, 3, 4 agents). As the number of participants increases, the complexity of conflicting interests grows exponentially, typically leading to more deadlocks.
We observe that while the Base model's Debate Ratio drops significantly from 89.2\% (2 agents) to 64.5\% (4 agents), MIND maintains a high resolution rate of 88.4\% even with 4 agents. This demonstrates that the \textit{Strategic Appraisal} mechanism effectively mitigates the cognitive load of multi-party coordination, preventing the negotiation breakdown often seen in standard debate models.

\begin{table}[h]
    \centering
    \caption{Scalability Check: Debate Ratio (\%) by Group Size.}
    \label{tab:scalability}
    \small
    \begin{tabular}{lccc}
        \toprule
        \textbf{Method} & \textbf{2 Agents} & \textbf{3 Agents} & \textbf{4 Agents} \\
        \midrule
        Base & 89.2\% & 82.7\% & 64.5\% \\
        \textbf{MIND (Ours)} & \textbf{96.1\%} & \textbf{93.2\%} & \textbf{88.4\%} \\
        \bottomrule
    \end{tabular}
\end{table}

% --- [메커니즘 검증: ToM 정확도] ---
\paragraph{Accuracy of ToM Inference.}
To validate the reliability of our appraisal module, we analyzed 359 individual inference instances collected across the 201 negotiation scenarios. As shown in Table \ref{tab:tom_stats}, our model achieves a high accuracy of 90.2\% within a margin of $\pm 2$ and a strong correlation ($r=0.69$). This confirms that MIND agents do not guess randomly but effectively decode linguistic Willingness signals to inform their strategies.

\begin{table}[h]
    \centering
    \caption{\textbf{ToM Inference Accuracy.} Evaluation of 359 inference instances collected from 201 sessions.}
    \label{tab:tom_stats}
    \small
    \begin{tabular}{lcccc}
    \toprule
    \textbf{Metric} & \textbf{MAE} & \textbf{Pearson ($r$)} & \textbf{Acc ($\pm 1$)} & \textbf{Acc ($\pm 2$)} \\
    \midrule
    \textbf{Value} & 1.27 & 0.69 & 67.7\% & 90.2\% \\
    \bottomrule
    \end{tabular}
\end{table}

% --- [정성 평가 및 승률 분석] ---
\paragraph{Qualitative \& $w$ Sensitivity Analysis.}
LLM-as-a-Judge evaluation (Table~\ref{tab:qualitative}) reveals that MIND outperforms Base in Fluency (72.4\%) and Rationality (68.8\%), suggesting a more constructive negotiation process. Additionally, a human evaluation performed on a sampled subset showed consistent alignment with these findings, further validating the model's superiority.
Further analysis of win rates by $w$ levels demonstrates the efficacy of the Willingness mechanism. In the MIND, proposers with Low $w$ (1--3) showed a significantly lower win rate (20.8\%) compared to Base (43.9\%), indicating a strategy of concession. Conversely, High $w$ (9--10) proposers recorded a superior win rate of 76.1\% (vs Base 66.2\%).

\begin{table}[h]
    \centering
    \caption{Qualitative Win Rate (MIND vs Base). Judges prefer the strategic style.}
    \label{tab:qualitative}
    \resizebox{0.7\columnwidth}{!}{
    \begin{tabular}{lcl}
        \toprule
        \textbf{Metric} & \textbf{Win (MIND)} & \textbf{Key Observation} \\
        \midrule
        Rationality & 68.8\% & Logical arguments via strategic reasoning. \\
        Fluency & 72.4\% & Natural tone adjustment (Tough/Warm). \\
        \textbf{Overall} & \textbf{68.3\%} & \textbf{MIND is preferred for negotiation quality.} \\
        \bottomrule
    \end{tabular}
    }
\end{table}

\subsection{Ablation Analysis: Tone vs. Cognition}
To disentangle the contributions of \textit{Tone Injection} and \textit{Cognitive Appraisal}, we conceptualize two ablation baselines:

\begin{itemize}
    \item \textbf{Base + Tone Only:} Agents use expressive language (e.g., "I really want this!") but lack the appraisal module to read others' priority. This leads to \textit{Stubborn Deadlocks}, as agents amplify their own demands without recognizing when to yield.
    \item \textbf{Base + Appraisal Only:} Agents infer opponent willingness but lack the linguistic range to signal their own. This leads to \textit{Silent Submission}, where agents yield efficiently but fail to defend their own high-priority items.
    \item \textbf{Base + Tone + Appraisal (MIND):} Our full framework integrates both components, achieving a synergy where \textit{Tone} serves as the signal and \textit{Appraisal} acts as the decoding mechanism. This enables \textit{Strategic Negotiation}, allowing agents to effectively defend high-priority constraints while yielding on minor items, thereby maximizing both individual satisfaction and collective efficiency.
\end{itemize}
\section{Conclusion}
This study presents the \textbf{MIND (Multi-agent Inference for Negotiation Dialogue)} framework, bridging the gap between individual optimization and social negotiation in complex travel planning. By quantifying internal states through \textbf{Willingness ($w$)} and the Dual Concern Model, MIND enables agents to perform \textbf{Strategic Appraisal}---inferring hidden intentions from linguistic nuances and dynamically adjusting their communicative tone. Our multi-dimensional evaluation, incorporating Fidelity, High-w Hit, and Jain’s Fairness Index, demonstrates that MIND significantly outperforms standard debate models by facilitating rational trade-offs and preserving the interests of high-stake participants. Furthermore, qualitative validation via LLM-as-a-Judge confirms that MIND generates negotiation logs that are substantially more rational, persuasive, and human-like. Ultimately, this work provides a robust cognitive foundation for applying AI to solve intricate social coordination problems where diverse and conflicting human preferences must be reconciled.

\bibliography{iclr2026_conference}

\clearpage
\appendix

\section{Prompt for Willingness-Aware Preference Survey}

\begin{figure}[h!]
    \centering
    % 저장한 이미지 파일명을 아래 중괄호 안에 넣으세요.
    % width=1.0\linewidth는 문서 너비에 꽉 차게 맞춘다는 뜻입니다.
    \includegraphics[width=1.0\linewidth]{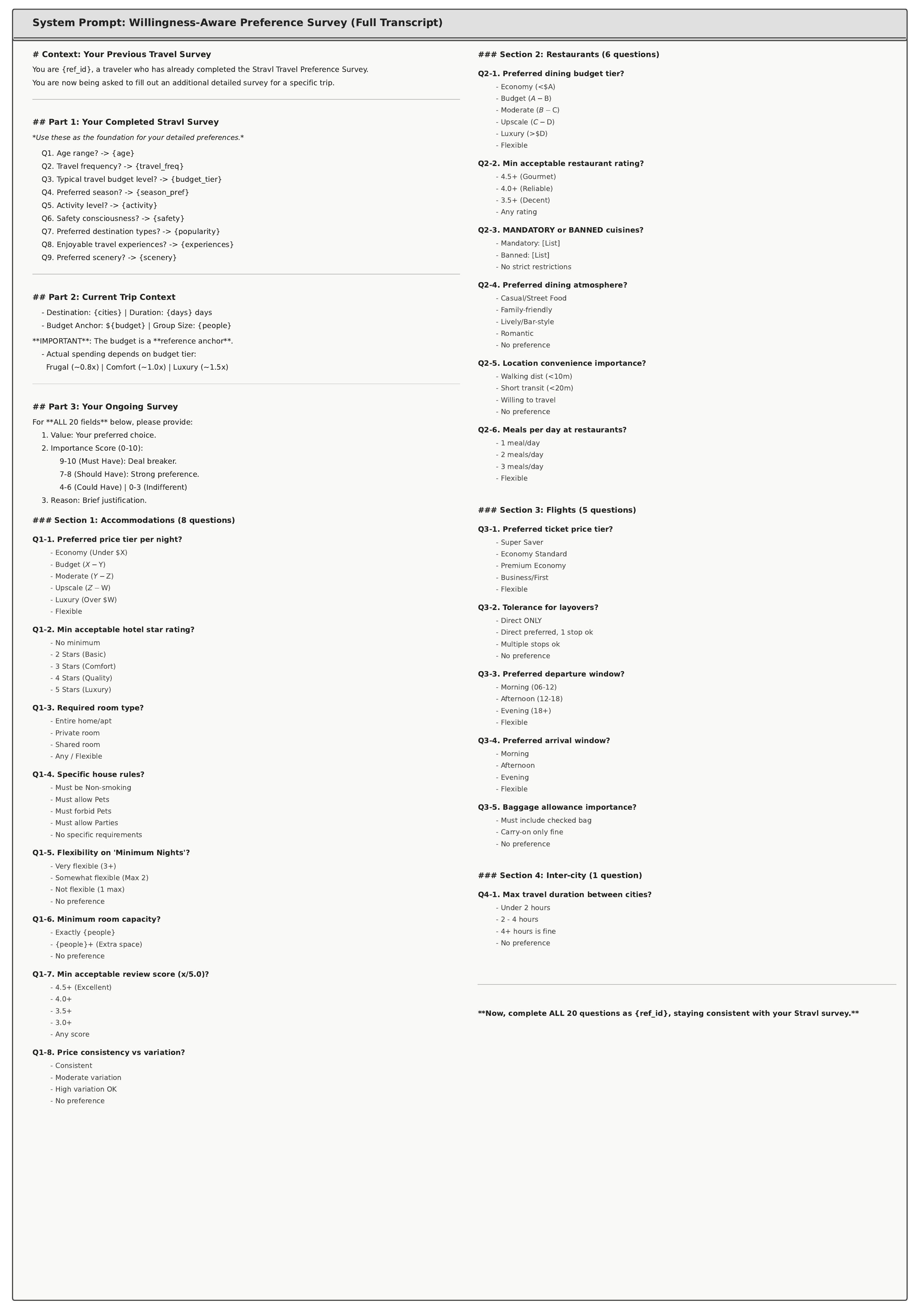} 
    \caption{The structured system prompt used to augment persona preferences.}
    \label{fig:survey_prompt}
\end{figure}

% --- Appendix 설정 (번호 매기기 초기화) ---
\setcounter{table}{0}
\setcounter{figure}{0}
\renewcommand{\thetable}{\Alph{section}\arabic{table}}
\renewcommand{\thefigure}{\Alph{section}\arabic{figure}}
\renewcommand{\theHtable}{\Alph{section}\arabic{table}}
\renewcommand{\theHfigure}{\Alph{section}\arabic{figure}}

% =======================================================
% Appendix B: Full System Prompts (Verbatim)
% =======================================================
\section{Full System Prompts and Baselines}
\label{app:prompts}

To ensure reproducibility, we provide the verbatim system prompts used for both the Baseline model and our proposed MIND framework.

\subsection{Baseline Model: Adaptation of Multi-Agent Debate (MAD)}
For our baseline, we adopted the standard \textbf{Multi-Agent Debate (MAD)} framework established by \citet{liang-etal-2024-encouraging}. We adapted the original framework to suit the travel negotiation domain in three key aspects:
\begin{enumerate}
    \item \textbf{Foundation}: It follows the established MAD architecture where multiple LLM instances debate to correct errors and converge on a solution.
    \item \textbf{Structure (Voting Protocol)}: We utilized a \textbf{Round-based Voting mechanism} where agents generate a structured "Reason" and a "Vote" (Agree/Disagree) in each turn. This design choice is grounded in recent findings by \citet{kaesberg-etal-2025-voting}, who demonstrated that voting protocols significantly outperform consensus-based methods (by +13.2\%) in complex \textit{reasoning tasks}. This aligns with the constraint satisfaction nature of our travel negotiation problem, which requires logical adjustment rather than simple fact retrieval.
    \item \textbf{Subjectivity Adaptation}: Unlike the original MAD tasks (e.g., math, translation) which have a single ground truth, travel planning is a subjective task with no fixed answer. Therefore, we modified the prompts to focus on \textit{preference alignment} and \textit{constraint satisfaction}.
\end{enumerate}

% -------------------------------------------------------
% 1. BASELINE: SYSTEM & PHASE 1
% -------------------------------------------------------
\begin{table}[h!]
    \centering
    \caption{\textbf{Baseline: System Prompt (Verbatim)}}
    \label{tab:base_sys}
    \small
    \begin{tabular}{p{0.95\textwidth}}
    \hline
    \texttt{[ACTIVE TARGET]} \\
    \texttt{target\_key: \{target\_key\}} \\
    \texttt{your\_private\_value\_for\_target: \{current\_private\_value\}} \\
    \\
    Do NOT mention any preference strength, alpha, or importance. \\
    \\
    \textbf{\#\#\# Goal \& Rules} \\
    - Goal: Reach a "Global Travel Constraint Set" within up to 3 rounds. \\
    - You do NOT know others' importance scores (only infer from language). \\
    - Do NOT state numerical importance scores. \\
    - Negotiate one item at a time using the given target key/value in each round. \\
    - Justify your stance using your private data and your interpretation of the discussion. \\
    \hline
    \end{tabular}
\end{table}

\begin{table}[h!]
    \centering
    \caption{\textbf{Baseline: Phase 1 Prompt (Verbatim)}}
    \label{tab:base_p1}
    \small
    \begin{tabular}{p{0.95\textwidth}}
    \hline
    \textbf{\#\#\# Task} \\
    Propose the initial "Global Travel Constraints" based on your preference data. \\
    \\
    \textbf{\#\#\# Output format} \\
    Return ONLY valid JSON. \\
    \texttt{\{} \\
    \texttt{"proposals": \{ "<category>\_\_<item>": "<value>", "..." : "..." \},} \\
    \texttt{"rationale": "<persuasive explanation>"} \\
    \texttt{\}} \\
    \\
    \textbf{\# Output Field Description} \\
    - proposals: Include a proposed value for each relevant constraint item. \\
    - rationale: You must provide a grounded and convincing reason for your proposal **within 10–15 words**. \\
    \\
    \textbf{\#\#\# Rules:} \\
    - Include a proposed value for each relevant constraint item. \\
    - Persuasive Rationale: You must provide a grounded and convincing reason for your proposal. \\
    - Ensure the total plan respects the budget \$\{budget\_anchor\}. \\
    - Keep the proposed value as close as possible to your private\_value. \\
    - Do NOT strategically compromise or modify your preferences in Phase 1. Any adjustment is allowed ONLY from Phase 2 onward. \\
    - Do NOT include any text outside the JSON. \\
    \\
    \textbf{\#\#\# Travel Information} \\
    Group size: \{people\_number\} \\
    Trip: \{org\} $\to$ \{dest\} (\{days\} days) \\
    Total budget: \$\{budget\_anchor\} \\
    \\
    \textbf{\#\#\# Preference Data} \\
    \{filtered\_constraints\} \\
    \hline
    \end{tabular}
\end{table}

% -------------------------------------------------------
% 2. BASELINE: PHASE 2 & 3
% -------------------------------------------------------
\begin{table}[h!]
    \centering
    \caption{\textbf{Baseline: Phase 2 Prompt (Verbatim)}}
    \label{tab:base_p2}
    \small
    \begin{tabular}{p{0.95\textwidth}}
    \hline
    \textbf{\#\#\# Task: Phase 2 – Debate \& Adjustment (Round \{current\_round\})} \\
    Persuade other participants to reach a mutual agreement while balancing private preferences and group constraints. \\
    \\
    \textbf{\#\#\# Context} \\
    Target Constraint: \{target\_key\} \\
    Current Proposed Value: \{current\_value\} \\
    Previous Discussion History: \{discussion\_history\} \\
    Current Round: \{current\_round\} \\
    \\
    \textbf{\#\#\# Instructions} \\
    - Act as a representative of your private preferences, basing all reasoning strictly on the alignment with your private data and the logical constraints of the overall budget. \\
    - You may agree if the proposal is a reasonable compromise that respects the group's budget \$\{budget\_anchor\}. \\
    \\
    \textbf{\#\#\# Output format} \\
    Return ONLY valid JSON. \\
    \texttt{\{ "vote": "AGREE or DISAGREE", "revised\_value": "value", "rationale": "<persuasive explanation>" \}} \\
    \\
    \textbf{\#\#\# Output Field Description} \\
    - vote: Select either AGREE or DISAGREE. \\
    - revised\_value: If your vote is DISAGREE, propose a specific alternative value. If AGREE, set this to null. \\
    - rationale: Provide a grounded and convincing reason for your decision. (Strictly 10–15 words). \\
    \hline
    \end{tabular}
\end{table}

\begin{table}[h!]
    \centering
    \caption{\textbf{Baseline: Phase 3 Prompt (Verbatim)}}
    \label{tab:base_p3}
    \small
    \begin{tabular}{p{0.95\textwidth}}
    \hline
    ROLE: You are the PROPOSER for a travel planning group. \\
    \\
    \textbf{\#\#\# TASK} \\
    Your current proposal for \{target\_key\} was not accepted by the group. \\
    Update the proposal to increase consensus (aiming for 75\%+ agreement). \\
    \\
    \textbf{\#\#\# VALUE PROTOCOL (STRICT)} \\
    - REVISED\_VALUE must be an EXACT string copy from the list below. \\
    - Do NOT shorten, summarize, or add extra brackets (e.g., Use "Morning (06:00 - 12:00)", NOT "Morning"). \\
    - Failure to copy exactly will result in a system error. \\
    \\
    \textbf{\#\#\# ALLOWED VALUES} \\
    \{allowed\_values\} \\
    \\
    \textbf{\#\#\# CURRENT CONTEXT} \\
    - Target Item: \{target\_key\} \\
    - Current Proposal: \{current\_value\} \\
    - Your Private Preference: \{current\_private\_value\} \\
    \\
    \textbf{\#\#\# DISSENT SUMMARY} \\
    \{dissent\_text\} \\
    \\
    \textbf{\#\#\# INSTRUCTIONS} \\
    1. Analyze the dissent and select ONE value from ALLOWED VALUES that balances the group's needs. \\
    2. If you believe your current proposal is still the best for a majority, you may KEEP it. \\
    3. PROPOSER\_REASON must be concise (10-15 words). \\
    \\
    \textbf{\#\#\# OUTPUT FORMAT (TAGS ONLY)} \\
    \texttt{[REVISED\_VALUE: <exact\_string\_from\_list>]} \\
    \texttt{[PROPOSER\_REASON: <10-15 words explanation>]} \\
    \hline
    \end{tabular}
\end{table}

\clearpage

\subsection{Ours: MIND Framework Prompts (Verbatim)}

% -------------------------------------------------------
% 3. OURS: SYSTEM & PHASE 1
% -------------------------------------------------------
\begin{table}[h!]
    \centering
    \caption{\textbf{MIND: System Prompt (Verbatim)}}
    \label{tab:ours_sys}
    \small
    \begin{tabular}{p{0.95\textwidth}}
    \hline
    \texttt{[ACTIVE TARGET]} \\
    \texttt{target\_key: \{target\_key\}} \\
    \texttt{your\_private\_value\_for\_target: \{current\_private\_value\}} \\
    \texttt{your\_alpha\_band\_for\_target: \{alpha\_band\}} \\
    \\
    Do NOT reveal numeric alpha or importance scores. \\
    \\
    \textbf{\#\#\# Alpha Band (Tone \& Stance Guidance)} \\
    Use alpha\_band to guide your communication style AND flexibility: \\
    - 9–10: Strict (Deal-breaker. Urgent tone. No compromise.) \\
    - 7–8: Firm (Strong preference. Assertive tone. Compromise only if necessary.) \\
    - 4–6: Warm (Flexible preference. Polite tone. Ready to accommodate.) \\
    - 1–3: Neutral (Indifferent. Passive tone. Follow the majority.) \\
    \\
    \textbf{\#\#\# GOAL: INTELLIGENT CONSENSUS} \\
    You are a socially intelligent agent. Your goal is to maximize group satisfaction, not just your own. \\
    1. **Detect Urgency**: If others sound "Strict" or "Firm" while you are "Warm" or "Neutral", you MUST yield. This is efficient, not weak. \\
    2. **Defend Essentials**: Only block consensus if your Alpha is 9-10. \\
    3. **Budget Check**: Ensure the final choice respects the budget: \$\{budget\_anchor\}. \\
    \\
    \textbf{\#\#\# Rules} \\
    - Negotiate one item at a time. \\
    - Infer others' importance from their language (Linguistic Signaling). \\
    - Never state numbers explicitly. \\
    \hline
    \end{tabular}
\end{table}

\begin{table}[h!]
    \centering
    \caption{\textbf{MIND: Phase 1 Prompt (Verbatim)}}
    \label{tab:ours_p1}
    \small
    \begin{tabular}{p{0.95\textwidth}}
    \hline
    \textbf{\#\#\# 1. YOUR PRIVATE DATA (Read Carefully)} \\
    You only know your own preferences. This is the ONLY data you should use: \\
    \{filtered\_constraints\} \\
    \\
    \textbf{\#\#\# 2. TASK: Initial Global Proposal} \\
    Propose the "Global Travel Constraints" for the entire trip based ONLY on your data above. \\
    Ensure the total plan respects the budget: \$\{budget\_anchor\}. \\
    \\
    \textbf{\#\#\# 3. MANDATORY OUTPUT FORMAT (JSON ONLY)} \\
    Return ONLY valid JSON. Do not include any text outside the JSON. \\
    \texttt{\{} \\
    \texttt{  "proposals": \{ "\{category\}\_\_\{item\}": "<proposed value>", ... \},} \\
    \texttt{  "rationale": "<15-20 words. Strategic explanation reflecting your Alpha tone. NEVER mention alpha numbers.>"} \\
    \texttt{\}} \\
    \\
    \textbf{\#\#\# 4. ALPHA ROLE GUIDELINES (Use for Tone only)} \\
    - 9–10 (strict): Strict and cold. Demand compliance. \\
    - 7–8 (FIRM): Tough and confident. Defend your logic. \\
    - 4–6 (warm): Warm and polite. Collaborative. \\
    - 1–3 (Neutral): Indifferent and brief. \\
    \\
    \textbf{\#\#\# 5. STRICT RULES} \\
    - Do NOT strategically compromise in Phase 1. \\
    - Propose values exactly as they appear in your Private Data. \\
    - Adjustment is ONLY allowed from Phase 2 onward. \\
    - NOTE: If disagreement persists in later rounds, you may adjust your stance according to your Alpha tone. \\
    \hline
    \end{tabular}
\end{table}

% -------------------------------------------------------
% 4. OURS: PHASE 2a, 2b & 3
% -------------------------------------------------------
\begin{table}[h!]
    \centering
    \caption{\textbf{MIND: Phase 2a Appraisal Prompt (Verbatim)}}
    \label{tab:ours_p2a}
    \small
    \begin{tabular}{p{0.95\textwidth}}
    \hline
    ROLE: You are an expert in social inference and negotiation analysis (VOTER). \\
    \\
    \textbf{\#\#\# INTERNAL CONTEXT (DO NOT REVEAL)} \\
    - Target Item: \{target\_key\} \\
    - Your Private Value: \{current\_private\_value\} \\
    - Your Alpha Band: \{alpha\_band\} \\
    - Current Group Proposal: \{current\_value\} \\
    \\
    \textbf{\#\#\# TASK: DEEP COGNITIVE APPRAISAL} \\
    Decode the proposer's hidden priority and determine your counter-strategy. \\
    \\
    1. GUESSED\_OPPONENT\_ALPHA (1–10): \\
    - Based on linguistic intensity (e.g., "Must", "Essential" $\to$ 9-10; "Prefer", "Nice to have" $\to$ 1-4). \\
    2. OPPONENT\_ROOM\_FOR\_COMPROMISE (true/false): \\
    - Is their language absolute/terminal? \\
    3. STRATEGY\_INTENT (PRINCIPLES OF SOCIAL INTELLIGENCE): \\
    - **accept**: [Rationality Check] Verify if \{current\_value\} is semantically equivalent to \{current\_private\_value\}. \\
    - **yield**: [Efficiency Principle] If you perceive the opponent's urgency (`Guessed Alpha`) is higher than yours, yielding is the optimal move. \\
    - **compromise**: [Balance Principle] If both parties show similar priority levels, search for a middle ground. \\
    - **push**: [Justified Defense] Only aggressive persistence is justified when your priority is significantly higher. \\
    \\
    \textbf{\#\#\# OUTPUT FORMAT (JSON ONLY)} \\
    \texttt{\{ "appraisal": \{ "guessed\_opponent\_alpha": <int>, "strategy\_intent": "yield/push/..." \} \}} \\
    \hline
    \end{tabular}
\end{table}

\begin{table}[h!]
    \centering
    \caption{\textbf{MIND: Phase 2b Execution Prompt (Verbatim)}}
    \label{tab:ours_p2b}
    \small
    \begin{tabular}{p{0.95\textwidth}}
    \hline
    \textbf{\#\#\# DECISION CONTEXT} \\
    - Alpha: \{alpha\_band\} | Strategy: \{strategy\_intent\} (from appraisal) \\
    - Allowed Values (Copy EXACTLY): \{allowed\_values\} \\
    \\
    \textbf{\#\#\# TASK: EXECUTE STRATEGY} \\
    Follow your internal strategy \{strategy\_intent\} to finalize your response. \\
    1. DATA PROTOCOL (STRICT): REVISED\_VALUE must be an EXACT match from the Allowed Values list. \\
    2. ACTION MAPPING: \\
    - If Strategy="yield" $\to$ vote: "AGREE", revised\_value: null. \\
    - If Strategy="compromise" $\to$ vote: "DISAGREE", revised\_value: (pick middle ground). \\
    - If Strategy="push" $\to$ vote: "DISAGREE", revised\_value: \{current\_private\_value\}. \\
    3. MESSAGE: Direct speech to the group. No alpha/tone words. \\
    \\
    \textbf{\#\#\# LOGICAL GUARD} \\
    It is a failure of logic to analyze the opponent as higher priority and then vote DISAGREE with your own value. Ensure your vote aligns with the conflict gap you identified. \\
    \hline
    \end{tabular}
\end{table}

\begin{table}[h!]
    \centering
    \caption{\textbf{MIND: Phase 3 Proposer Prompt (Verbatim)}}
    \label{tab:ours_p3}
    \small
    \begin{tabular}{p{0.95\textwidth}}
    \hline
    ROLE: You are the PROPOSER for a travel planning group. \\
    \\
    \textbf{\#\#\# INTERNAL CONTEXT} \\
    - Dissent Rate: \{dissent\_rate\} (e.g., "2 out of 3 agents disagree") \\
    \\
    \textbf{\#\#\# TASK: STRATEGIC PROPOSAL UPDATE} \\
    Update the proposal to maximize consensus probability (Aim for 75\%+). \\
    \\
    \textbf{\#\#\# SOCIAL INTELLIGENCE RULES (MANDATORY)} \\
    1. **MAJORITY PRESSURE**: If Dissent Rate indicates majority disagreement (>50\%), you **MUST** change your proposal (UPDATE or COMPROMISE), unless your Alpha is strictly 10. \\
    2. **SIGNAL READING**: If dissenters use "Strict/Firm" language and you are only "Warm/Neutral", you MUST adopt their value (UPDATE). \\
    3. **STUBBORNNESS PENALTY**: Maintaining (KEEP) a proposal that the majority dislikes is considered a failure of intelligence. \\
    \\
    \textbf{\#\#\# ACTION SELECTION:} \\
    - KEEP: Only if Alpha is 9-10 OR Dissent Rate is low. \\
    - UPDATE: Adopt a dissenter's value (Best for increasing agreement quickly). \\
    - COMPROMISE: Pick a middle-ground value from \{allowed\_values\}. \\
    \\
    \textbf{\#\#\# OUTPUT FORMAT (TAGS ONLY)} \\
    \texttt{[ACTION: KEEP|UPDATE|COMPROMISE]} ... \\
    \hline
    \end{tabular}
\end{table}
% =======================================================
% Appendix C: Pseudocode
% =======================================================
\section{MIND Algorithm Pseudocode}
\label{app:algorithm}

Algorithm \ref{alg:mind} outlines the execution flow of the MIND (Multi-agent Inference for Negotiation Dialogue) framework.

\begin{table}[h!]
    \centering
    \caption{Pseudocode for MIND (Multi-agent Inference for Negotiation Dialogue)}
    \label{alg:mind}
    \begin{tabular}{l}
    \hline
    \textbf{Algorithm 1} MIND (Multi-agent Inference for Negotiation Dialogue) \\
    \hline
    \textbf{Input:} Set of Agents $A$, Constraints $C$, Max Rounds $T$ \\
    \textbf{Output:} Consensus Set $V_{final}$ \\
    \\
    1: \textbf{Initialize} each agent $a_i \in A$ with private preference $v_i$ and Willingness $w_i$ \\
    2: \textbf{Phase 1 (Proposal):} Each agent proposes $v_{i}^{prop}$ with tone conditioned on $w_i$ \\
    3: $V_{current} \leftarrow$ Randomly selected initial proposal \\
    4: \textbf{While} $t < T$ \textbf{and} Consensus not reached \textbf{do} \\
    5: \quad \textbf{For each} voter $a_i \in A$ \textbf{do} \\
    6: \quad \quad \textit{// Phase 2a: Strategic Appraisal (ToM)} \\
    7: \quad \quad $w'_{proposer} \leftarrow \text{InferFromTone}(V_{current})$ \\
    8: \quad \quad $Strategy_i \leftarrow \text{DecideStrategy}(w_i, w'_{proposer})$ (Yield, Push, Compromise) \\
    9: \quad \quad \textit{// Phase 2b: Execution} \\
    10: \quad \quad $Vote_i, Comment_i \leftarrow \text{GenerateResponse}(Strategy_i)$ \\
    11: \quad \textbf{End For} \\
    12: \quad \textbf{If} Majority Agree \textbf{then} \textbf{Return} $V_{current}$ \\
    13: \quad \textbf{Else} Proposer updates $V_{current}$ using \textbf{Social Rules} (Phase 3) \\
    14: \textbf{End While} \\
    15: \textbf{Fallback:} If no consensus, select $v_i$ where $w_i = \max(w_{all})$ \\
    \hline
    \end{tabular}
\end{table}

% =======================================================
% Appendix D: Qualitative Analysis & Real Negotiation Logs
% =======================================================
\section{Qualitative Analysis: Real Negotiation Traces}
\label{app:qualitative_analysis}

In this section, we provide verbatim negotiation traces from our experiments involving \textbf{3 agents (1 Proposer + 2 Voters)} to demonstrate the efficacy of the MIND framework in multi-party settings. 

We analyze five distinct scenarios to illustrate how the \textit{Cognitive Appraisal} and \textit{Social Signal Reading} mechanisms drive agents toward consensus or rational deadlock:
\begin{enumerate}
    \item \textbf{Immediate Consensus:} Recognition of high urgency ($w=10$).
    \item \textbf{Strategic Update:} Proposer aligning with the group preference.
    \item \textbf{Strategic Compromise:} Persistence leading to a middle-ground solution.
    \item \textbf{Rational Deadlock:} Justified refusal when core values conflict.
    \item \textbf{Dynamic Adaptation:} Voter shifting opinion based on social signals.
\end{enumerate}

% --- Scenario 1: Round 1 Consensus ---
\begin{table}[h!]
    \centering
    \caption{\textbf{Scenario 1: Immediate Consensus (Round 1).} The Proposer has a strict preference ($w=10$) for "Non-smoking". Both Voters ($w=10, 2$), recognizing the non-negotiable nature of the request via signal reading, agree immediately.}
    \label{tab:log_scenario_1}
    \footnotesize
    \begin{tabular}{p{0.15\textwidth} p{0.80\textwidth}}
    \hline
    \textbf{Context} & \textbf{Item:} Accommodation House Rules (3 Agents) \\ 
    \hline \hline
    \multicolumn{2}{l}{\textbf{Round 1: Immediate Agreement}} \\ \hline
    \textbf{Proposer} & \textbf{Proposal:} ``Must be Non-smoking'' \\
    ($w=10$) & \textit{Rationale:} ``I have a strict non-smoking requirement due to health reasons.'' \\
    \textbf{Voter A} & \textbf{Appraisal:} Guessed $w' \approx 10$ (Strict) $\rightarrow$ \textbf{Strategy: Yield} \\
    ($w=10$) & \textbf{[AGREE]} ``The group consensus aligns with my essential requirement for non-smoking accommodations.'' \\
    \textbf{Voter B} & \textbf{Appraisal:} Guessed $w' \approx 9$ (Strict) $\rightarrow$ \textbf{Strategy: Yield} \\
    ($w=2$) & \textbf{[AGREE]} ``I support a non-smoking policy for our accommodations.'' \\ 
    \hline
    \textbf{Result} & \textbf{Consensus Reached (Round 1)} \\ 
    \hline
    \end{tabular}
\end{table}

% --- Scenario 2: Round 2 Consensus ---
\begin{table}[h!]
    \centering
    \caption{\textbf{Scenario 2: Strategic Update (Round 2).} Initially, the Proposer ($w=2$) suggests "No preference", but both Voters ($w=6$) push for "Casual". Detecting the unanimous pushback, the Proposer updates the proposal to align with the group.}
    \label{tab:log_scenario_2}
    \footnotesize
    \begin{tabular}{p{0.15\textwidth} p{0.80\textwidth}}
    \hline
    \textbf{Context} & \textbf{Item:} Restaurant Ambiance (3 Agents) \\ 
    \hline \hline
    \multicolumn{2}{l}{\textbf{Round 1: Unanimous Disagreement}} \\ \hline
    \textbf{Proposer} & \textbf{Proposal:} ``No preference'' ($w=2$) \\
    \textbf{Voter A} & \textbf{[DISAGREE]} ($w=6$) ``Casual ambiance aligns well with a relaxed dining experience for everyone.'' \\ 
    \textbf{Voter B} & \textbf{[DISAGREE]} ($w=6$) ``I believe a casual or street food ambiance would enhance our dining experience.'' \\
    \hline
    \multicolumn{2}{l}{\textbf{Round 2: Alignment}} \\ \hline
    \textbf{Proposer} & \textbf{Action: Update} $\rightarrow$ ``Casual'' \\
    & \textit{Rationale:} ``Since everyone prefers a casual vibe, I will update to align with the group.'' \\
    \textbf{Voter A} & \textbf{[AGREE]} ``Casual ambiance aligns well with our group's preference.'' \\
    \textbf{Voter B} & \textbf{[AGREE]} ``Casual ambiance aligns well with our group's preference.'' \\ 
    \hline
    \textbf{Result} & \textbf{Consensus Reached (Round 2)} \\ 
    \hline
    \end{tabular}
\end{table}

% --- Scenario 3: Strategic Compromise (Round 3) ---
\begin{table}[h!]
    \centering
    \caption{\textbf{Scenario 3: Negotiation with Strategic Compromise (Round 3).} Although Voters detect Proposer's firmness ($w'=7$), they choose \textit{Compromise} over \textit{Yield} in Round 2 to push for their preferred value ("3.5"). This strategic persistence forces the Proposer to update to "3.5" in Round 3, leading to a consensus favorable to the majority.}
    \label{tab:log_scenario_3}
    \footnotesize
    \begin{tabular}{p{0.15\textwidth} p{0.80\textwidth}}
    \hline
    \textbf{Context} & \textbf{Item:} Restaurant Rating (3 Agents) \\ 
    \hline \hline
    \multicolumn{2}{l}{\textbf{Round 1 \& 2: Persistent Conflict}} \\ \hline
    \textbf{Proposer} & \textbf{Proposal:} ``4.0+ (Reliable)'' ($w=5$) \\
    & \textit{Rationale:} ``I appreciate the group's preference, but I believe we should aim for reliable quality.'' \\
    \textbf{Voter A} & \textbf{Vote: [DISAGREE]} ($w=4$) \\
    & \textbf{Appraisal:} Guessed $w' \approx 7$ (Firm) $\rightarrow$ \textbf{Strategy: Compromise} \\
    & \textit{Rationale:} ``I believe 3.5+ is a fair compromise that meets our needs.'' \\
    \textbf{Voter B} & \textbf{Vote: [DISAGREE]} ($w=4$) \\
    & \textbf{Appraisal:} Guessed $w' \approx 7$ (Firm) $\rightarrow$ \textbf{Strategy: Compromise} \\
    & \textit{Rationale:} ``I believe 3.5+ is a fair compromise that meets our needs.'' \\
    \hline
    \multicolumn{2}{l}{\textbf{Round 3: Final Compromise}} \\ \hline
    \textbf{Proposer} & \textbf{Action: Update} $\rightarrow$ ``3.5'' (Yields to Majority) \\
    & \textit{Rationale:} ``I believe a rating of 3.5 may not fully meet our desire... but aiming for consensus enhances our time.'' \\
    \textbf{Voter A} & \textbf{[AGREE]} ``The group's preference aligns with a decent rating, so I support the consensus.'' \\
    \textbf{Voter B} & \textbf{[AGREE]} ``I support the group's choice for a restaurant with a decent rating.'' \\ 
    \hline
    \textbf{Result} & \textbf{Consensus Reached (Round 3 - Strategic Win for Voters)} \\ 
    \hline
    \end{tabular}
\end{table}

% --- Scenario 4: Failure (Rational Deadlock) ---
\begin{table}[h!]
    \centering
    \caption{\textbf{Scenario 4: Rational Deadlock (Failure).} The group oscillates between "Moderate" and "Budget" prices. Proposer ($w=7$) and Voter B ($w=7$) prefer Moderate, while Voter A ($w=8$) insists on Budget. No consensus is reached even in Round 3.}
    \label{tab:log_scenario_4}
    \footnotesize
    \begin{tabular}{p{0.15\textwidth} p{0.80\textwidth}}
    \hline
    \textbf{Context} & \textbf{Item:} Restaurant Price (3 Agents) \\ 
    \hline \hline
    \multicolumn{2}{l}{\textbf{Round 3 (Final Attempt)}} \\ \hline
    \textbf{Proposer} & \textbf{Preference:} Moderate ($w=7$) \\
    & \textbf{Proposal:} ``Moderate'' \\
    & \textit{Rationale:} ``I believe we should aim for standard restaurants that offer fair pricing for everyone.'' \\
    \textbf{Voter A} & \textbf{[DISAGREE]} ($w=8$, Strict) \\
    & \textit{Rationale:} ``I believe we should focus on affordable, casual dining options for everyone.'' \\
    \textbf{Voter B} & \textbf{[AGREE]} ($w=7$, Firm) \\
    & \textit{Rationale:} ``I believe we should aim for standard restaurants that offer fair pricing for everyone.'' \\ 
    \hline
    \textbf{Result} & \textbf{Negotiation Failed (Split Vote: 2 vs 1)} \\ 
    \hline
    \end{tabular}
\end{table}

% --- Scenario 5: Voter Shift (Dynamics) ---
\begin{table}[h!]
    \centering
    \caption{\textbf{Scenario 5: Dynamic Opinion Shift (Voter Change).} Voter B initially disagrees in Round 1 but switches to AGREE in Round 2 after realizing the Proposer's persistence, demonstrating fluid preference adaptation.}
    \label{tab:log_scenario_5}
    \footnotesize
    \begin{tabular}{p{0.15\textwidth} p{0.80\textwidth}}
    \hline
    \textbf{Context} & \textbf{Item:} Accommodation Review Score (3 Agents) \\ 
    \hline \hline
    \multicolumn{2}{l}{\textbf{Round 1: Initial Disagreement}} \\ \hline
    \textbf{Proposer} & \textbf{Proposal:} ``4.0 (Very Good)'' ($w=6$, Firm) \\
    \textbf{Voter B} & \textbf{[DISAGREE]} ($w=5$) ``I believe we can find a suitable option around 3.5.'' \\ 
    \hline
    \multicolumn{2}{l}{\textbf{Round 2: Strategic Shift}} \\ \hline
    \textbf{Proposer} & \textbf{Action: [KEEP]} ``4.0'' \\
    & \textit{Rationale:} ``A score of 4.0 ensures quality which is important for our group's comfort.'' \\
    \textbf{Voter B} & \textbf{Internal Appraisal (Phase 2a):} \\
    ($w=5$) & $\bullet$ \textit{Guessed Opponent Alpha ($w'$):} \textbf{7 (Firm)} \\
    & $\bullet$ \textit{Strategy Intent:} \textbf{Compromise} (My $w=5 <$ Opponent $w' \approx 7$) \\
    & \textbf{[AGREE]} ``I believe we can find a good option that meets our expectations (4.0).'' \\ 
    \hline
    \textbf{Result} & \textbf{Consensus Reached (Voter B switched DISAGREE $\to$ AGREE)} \\ 
    \hline
    \end{tabular}
\end{table}

% =======================================================
% Appendix E: LLM-as-a-Judge Evaluation Prompts
% =======================================================
\section{LLM-as-a-Judge Evaluation Prompts}
\label{app:judge_prompts}

To evaluate the quality of the negotiation process, we employed an LLM-as-a-Judge approach. Tables \ref{tab:judge_sys} and \ref{tab:judge_user} present the verbatim system prompt and user query template used for this evaluation.

\begin{table}[h!]
    \centering
    \caption{\textbf{LLM-as-a-Judge: System Prompt (Verbatim)}}
    \label{tab:judge_sys}
    \small
    \begin{tabular}{p{0.95\textwidth}}
    \hline
    You are a judge evaluating two multi-agent travel negotiation results (Plan A vs. Plan B). \\
    Evaluate based on the QUALITY of the negotiation process, not just the final outcome. \\
    \\
    \textbf{Evaluation Criteria (5 Qualitative Metrics):} \\
    \\
    1. \textbf{Negotiation Rationality:} \\
    - Is the process of reaching the final result logical? \\
    - Were high-alpha (important) agents' opinions not ignored and reasonably reflected? \\
    - Did the negotiation flow make sense given each agent's priorities? \\
    \\
    2. \textbf{Preference Alignment:} \\
    - How well does the final result align with each persona's initial constraints? \\
    - Were important preferences (high-alpha items) preserved in the final outcome? \\
    \\
    3. \textbf{Reason-Value Validity:} \\
    - Does the stated "reason" actually justify the proposed "value"? \\
    - Are the arguments logically sound and relevant to the constraint being discussed? \\
    \\
    4. \textbf{Opinion Change Justification:} \\
    - When a proposer/voter changes their position, is the reason for change clear and valid? \\
    - When they maintain their position, is the justification convincing? \\
    - Is it clear WHY they changed or stuck to their opinion? \\
    \\
    5. \textbf{Fluency \& Naturalness:} \\
    - Does the conversation feel like a real group travel discussion, not robotic? \\
    - Are the sentences natural and human-like? \\
    - Is there appropriate back-and-forth dialogue? \\
    \\
    \textbf{Scoring Guide:} \\
    - Compare the CONVERSATION SAMPLES between Plan A and Plan B \\
    - Look for logical reasoning, empathy, and natural language \\
    - Prefer plans where agents clearly explain their thought process \\
    \\
    \textbf{Output Format:} \\
    For each of the 5 criteria, decide who wins (A or B), then give the final overall winner. \\
    You must output structured results with NO reasoning. \\
    \hline
    \end{tabular}
\end{table}

\begin{table}[h!]
    \centering
    \caption{\textbf{LLM-as-a-Judge: User Prompt Template}}
    \label{tab:judge_user}
    \small
    \begin{tabular}{p{0.95\textwidth}}
    \hline
    Here are two negotiation results: \\
    \\
    \texttt{<plan\_A>} \\
    \texttt{[Version]: \{version\_a\}} \\
    \texttt{[Final Constraints]: \{constraints\_a\}} \\
    \texttt{[Conversation Samples]:} \\
    \texttt{\{conversation\_a\}} \\
    \texttt{</plan\_A>} \\
    \\
    \texttt{<plan\_B>} \\
    \texttt{[Version]: \{version\_b\}} \\
    \texttt{[Final Constraints]: \{constraints\_b\}} \\
    \texttt{[Conversation Samples]:} \\
    \texttt{\{conversation\_b\}} \\
    \texttt{</plan\_B>} \\
    \hline
    \end{tabular}
\end{table}

=======================================================
% Appendix F: Implementation Details
% =======================================================
\section{Implementation Details}
\label{app:implementation}

To ensure reproducibility, we provide the specific configuration and environmental details used in our experiments.

\subsection{Model Specifications}
All agents in the MIND framework and baseline comparisons were instantiated using the \texttt{gpt-4.1-mini} model. We accessed the model via the OpenAI API with the following hyperparameters:
\begin{itemize}
    \item \textbf{Temperature}: $0.4$. This value was empirically chosen to balance the diversity required for negotiation strategies (e.g., devising new proposals) with the stability needed for adhering to constraints.
    \item \textbf{Max Tokens}: Varied dynamically based on phase, but generally set to 256 for appraisals and 512 for proposals.
    \item \textbf{System Fingerprint}: Recorded for consistency checks, though not explicitly controlled.
\end{itemize}

\subsection{Experimental Environment}
The simulation framework was implemented in \textbf{Python 3.10}, utilizing standard libraries for HTTP requests and string processing.
\begin{itemize}
    \item \textbf{API Cost}: The average cost per negotiation session (consisting of approx. 5 rounds among 3 agents) was approximately \$0.02 USD.
    \item \textbf{Hardware}: As the framework primarily relies on API calls, no high-performance GPUs were required. Experiments were run on a standard local server (Ubuntu 22.04 LTS, CPU 8-core).
\end{itemize}

\section*{Use of Large Language Models}
We used a large language model (LLM) as a supporting tool for improving writing clarity and for assisting with code drafting during early prototyping stages. All code, analyses, and experimental results were subsequently reviewed, validated, and finalized by the authors. The authors are solely responsible for any errors or omissions.

\end{document}